\PassOptionsToPackage{table}{xcolor}
\documentclass[11pt]{article}

\usepackage[final]{acl}

\usepackage{times}
\usepackage{latexsym}
\usepackage{algorithm}
\usepackage{algpseudocode}
\usepackage{amsmath}
\usepackage{amssymb}
\usepackage{booktabs}
\usepackage{tabularx}
\usepackage{multirow}
\usepackage{array}
\usepackage{caption}
\usepackage{placeins}
\usepackage[T1]{fontenc}

\usepackage[utf8]{inputenc}

\usepackage{microtype}

\usepackage{inconsolata}

\usepackage{graphicx}

\usepackage{pifont}
\definecolor{myrowblue}{RGB}{234,244,255}
\definecolor{myrowblue_1}{RGB}{214,234,248}

\usepackage{graphicx}
\title{ \raisebox{-0.35em}{\includegraphics[height=2em]{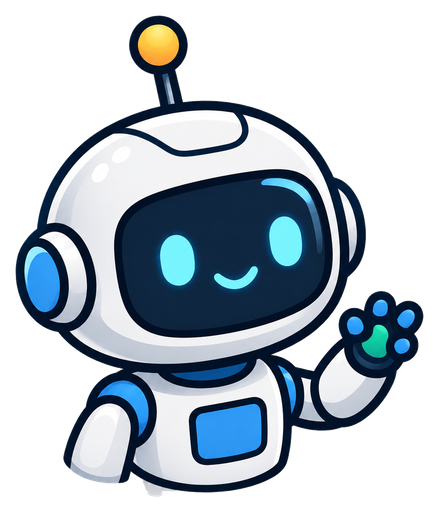}}\hspace{0.25em} Skill-Guided Continuation Distillation for GUI Agents }

\usepackage{graphicx}
\usepackage{hyperref}
\usepackage[most]{tcolorbox}
\usepackage{enumitem}

\author{ \textbf{Zhimin Fan$^{1,*}$ \quad Hongwei Yu$^{1,2,*}$ \quad Yeqing Shen$^{1,\dagger}$ \quad Haolong Yan$^{1}$} \\ \textbf{Guozhen Peng$^{1}$ \quad Tianhao Peng$^{4}$ \quad Yudong Zhang$^{3}$ \quad Xiaowen Zhang$^{2,\ddagger}$} \\ \textbf{Kaijun Tan$^{1}$ \quad Zheng Ge$^{1}$ \quad Xiangyu Zhang$^{1}$ \quad Daxin Jiang$^{1}$} \\ $^{1}$StepFun \quad $^{2}$University of Science and Technology Beijing \quad $^{3}$Tsinghua University \\ $^{4}$Nanyang Technological University \\ $^{*}$Equal contribution. \quad $^{\dagger}$Project lead. \quad $^{\ddagger}$Corresponding author. }

\begin{document}
\maketitle
\begin{abstract}



Improving GUI agents typically relies on behavior cloning on expert trajectories.
However, as the current policy deviates from the expert policy, it inevitably encounters policy-induced off-trajectory states during closed-loop execution, i.e., states that fall outside the expert trajectories.
Since expert trajectories provide no demonstrations for these unseen states, such states receive no effective supervision, leaving the policy unable to select the correct action.
To close this supervision gap, we propose Skill-Guided Continuation Distillation (SGCD), an iterative self-improvement framework. SGCD first runs the plain policy without skill guidance for a few steps to reach realistic off-trajectory states. From these states, a skill-guided policy then completes the task and produces successful continuations, which are mixed with expert trajectories to supply supervision over policy-induced off-trajectory states. The skills are extracted from both successful and failed rollouts, consisting of Continuation Plans, Critical Targets, Failure Traps, and Success Criteria.
On OSWorld-Verified, SGCD improves the success rate of three base models from the low-30\% range to over 50\%, demonstrating its effectiveness and generality.


\end{abstract}

\section{Introduction}

Built on recent vision-language foundation models \cite{google2025gemini3, bai2025qwen3, yan2025step, openai2026gpt54,anthropic2025claude4, hurst2024gpt}, GUI agents perceive screen observations and predict actions to operate desktop, web, and mobile interfaces in a closed loop, supporting open-ended computer tasks such as document editing, software operation, and web navigation. Such agents are typically trained by supervised fine-tuning (SFT) over trajectory data, which adapts the underlying vision-language models to GUI-specific observations, action spaces, and interaction protocols.

Existing end-to-end GUI agents are trained on human or synthetic expert trajectories \cite{qin2025ui,wang2026opencua,yan2025step,xu2026mobile,xue2026evocua}, teaching task-specific behaviors, action formats, and procedural knowledge. Self-improvement methods further expand the training pool by converting model-generated rollouts into supervision through filtered self-training, sandbox-based reinforcement learning, or experience-driven knowledge refinement \cite{yan2025step,wu2026litegui,lai2025computerrl,zhang2025ui,lin2026ui,wang2026off}. Despite differences in data sources, these approaches share a common supervision paradigm: behavior cloning on successful expert trajectories, where the policy is trained to imitate the action taken at each visited expert state.
While such supervision is strong on the expert state distribution, the discrepancy between the expert policy and the current policy inevitably drives the current policy into states that fall outside the expert trajectories~\cite{lauffer2025imitation,ross2011reduction}.
We refer to these states as \textbf{policy-induced off-trajectory states}.
Expert trajectories provide no effective supervision for these states, leaving the policy unable to predict the correct action. Reinforcement learning is explored as an alternative source of supervision for such states \cite{lai2025computerrl, li2025efficient}, but rollouts from the current policy rarely produce correct actions, yielding sparse reward signals and inefficient training \cite{zeng2025reinforcing, wang2026information}.

\begin{figure*}[t]
    \centering
    \includegraphics[width=1.0\linewidth]{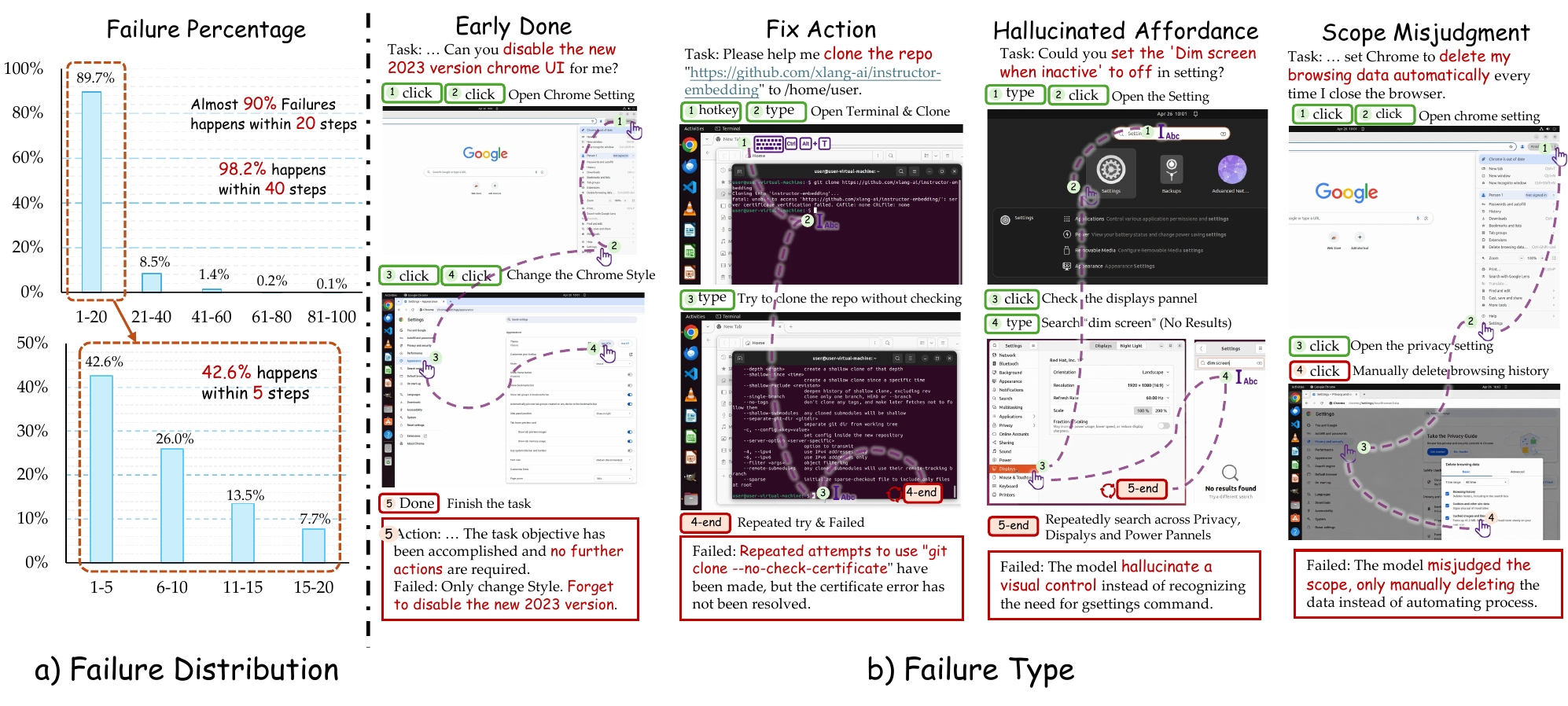}
    \caption{Failure analysis of GUI agents. Left: failures are concentrated in early execution. Right: representative recurring failure patterns, including early done, fix action, hallucinated affordance, and scope misjudgment.}
    \label{fig:failure-analysis}
\end{figure*}


Unlike single-step prediction, where each output is independent, end-to-end tasks involve long sequences of actions executed in a closed loop. An early off-trajectory action thus propagates through subsequent interactions and drives the agent into states increasingly far from the expert trajectories~\cite{chen2025training}. Importantly, such states are not arbitrary perturbations of expert states but reflect systematic biases of the learned policy, which tends to repeat a small set of erroneous behaviors. 
We refer to this distributional and supervisory mismatch as the off-trajectory supervision deficit.


Closing this supervision deficit is particularly challenging in GUI domains. Reaching realistic off-trajectory states requires actually executing actions in the environment, making such states costly to revisit, reproduce, or reset.
Existing methods rely on hand-crafted rules to select important off-trajectory states \cite{lin2026ui}, but such heuristics introduce selection bias and yield only sparse coverage of the states the current policy traverses.
Moreover, obtaining successful continuations is difficult, as the current policy seldom completes the task from such states. An effective method for supplying such supervision should therefore \ding{172} \textbf{expose the policy to realistic off-trajectory states} and \ding{173} \textbf{obtain successful continuations} from such states.


To meet both objectives, we propose Skill-Guided Continuation Distillation (SGCD), an iterative self-improvement framework. SGCD rolls out the plain policy (i.e., the current policy without skill guidance) to reach realistic off-trajectory states, then invokes a skill-guided policy to complete the task and produce successful continuations from those states. Mixing verified continuations with expert trajectories for training supplies additional supervision over policy-induced off-trajectory states. Concretely, each objective is realized as follows.


For objective \ding{172}, we first analyze where GUI failures occur during trajectories.
As shown in Fig.~\ref{fig:failure-analysis}(a), failures are strongly concentrated in early execution, suggesting that off-trajectory deviations are often induced early and can lead to erroneous subsequent actions and eventual task failure. 
Accordingly, we induce off-trajectory states from the early execution of the plain policy, where such deviations naturally arise.
For each task, the plain policy interacts with the GUI for $k$ steps to instantiate realistic off-trajectory states.
By sweeping $k$ over a range of values, SGCD avoids hand-crafted state-selection heuristics and densely covers the off-trajectory states the current policy actually traverses.
This aligns the supervision distribution with the states the policy encounters at deployment, directly mitigating the distributional shift inherent in behavior cloning.


For objective \ding{173}, we analyze the structured failure patterns induced by the learned policy. As illustrated in Fig.~\ref{fig:failure-analysis}(b), policy failures exhibit recurring error tendencies rather than isolated accidental mistakes~\cite{wanyan2026look,lu2025agentrewardbench}. From successful and failed rollouts, we extract \emph{off-trajectory continuation skills} (Continuation Plans, Critical Targets, Failure Traps, and Success Criteria), which guide a skill-guided policy to roll out from each off-trajectory state and produce verified successful continuations for training.
By fulfilling these two objectives, SGCD synthesizes effective supervision over policy-induced off-trajectory states, closing the supervision deficit inherent in behavior cloning on expert trajectories.

We evaluate our method on OSWorld-Verified across three vision-language models: Qwen3-VL-8B, Qwen3-VL-30B-A3B, and STEP3-VL-10B. Across all three models, SGCD consistently improves performance from the low-30\% range to over 50\%. The main contributions are as follows:




\begin{itemize}
    \item We identify the \textbf{off-trajectory supervision deficit}, where agents trained on successful demonstrations lack supervision for policy-induced off-trajectory states, and show that failures are concentrated in early stage.

    \item We propose \textbf{Skill-Guided Continuation Distillation}, which uses off-trajectory continuation skills to collect successful continuations from off-trajectory states and mitigate the expert-state bias.

    \item We validate SGCD on OSWorld-Verified across three base models, with success rates improving from the low-30\% range to over 50\%, demonstrating its generality.
\end{itemize}

\begin{figure*}[t]
    \centering
    \includegraphics[width=0.98\linewidth]{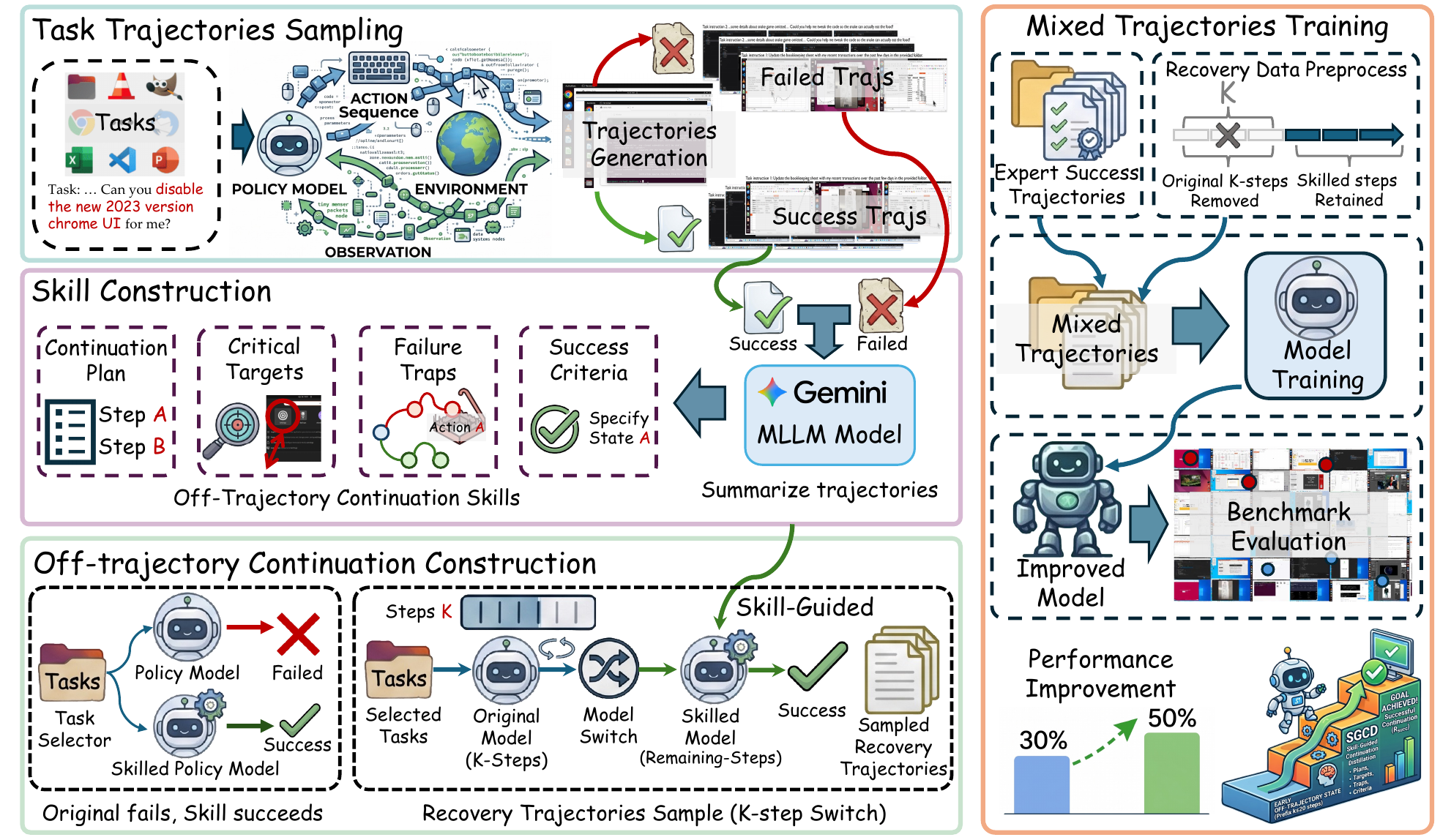}
\caption{
Overview of SGCD. 
(1) \textbf{Task Trajectories Sampling}: collect successful and failed plain-policy trajectories.
(2) \textbf{Skill Construction}: extract off-trajectory continuation skills from trajectory evidence.
(3) \textbf{Off-trajectory Continuation Construction}: use $k$-step handoff to collect skill-guided successful continuations.
(4) \textbf{Mixed Trajectories Training}: train the plain policy with expert and verified continuation trajectories.
}
    \label{fig:method}
\end{figure*}

\section{Related Work}

\paragraph{GUI Agents.}
Recent GUI agents advance on web, mobile, and desktop benchmarks~\cite{deng2023mind2web,zhou2024webarena,rawles2025androidworld,xie2024osworld} along several directions. UI-TARS~\cite{qin2025ui} unifies perception, reasoning, and action generation through large-scale GUI-specific pretraining. OpenCUA~\cite{wang2026opencua} scales human-annotated desktop trajectories with reflective state-action conversion to support open-ended tasks. SeeClick~\cite{cheng2024seeclick} and UGround~\cite{qian2025uground} target element grounding via screen-localized pretraining for accurate UI localization. EvoCUA~\cite{xue2026evocua} and LiteGUI~\cite{wu2026litegui} automatically synthesize task and trajectory data to continuously update the policy. These works establish strong foundational capabilities for GUI agents and demonstrate effective data synthesis pipelines. Building on this foundation, SGCD proposes an iterative self-improvement approach that further addresses the lack of expert supervision for policy-induced off-trajectory states.


\paragraph{Self-Improvement.}
Self-improvement methods seek to enhance agent performance by leveraging the agent's own interaction experience as a training signal.
Reflexion~\cite{shinn2023reflexion} and Self-Refine~\cite{madaan2023self} use inference-time verbal feedback to revise outputs. Recent GUI pipelines convert model-generated rollouts into supervision through filtered self-training~\cite{yan2025step,wu2026litegui}, sandbox-based reinforcement learning~\cite{lai2025computerrl}, experience-driven knowledge refinement~\cite{zhang2025ui,lin2026ui}, or policy-aligned experience assimilation~\cite{wang2026off}.
SGCD follows this paradigm and specifically targets policy-induced off-trajectory states, synthesizing skill-guided continuation supervision from the states the current policy actually traverses.
Detailed related work is shown in Appendix~\ref{sec:extended-related-work}.

\section{Preliminaries}
\label{sec:problem-setup}

We consider a distribution of executable GUI training tasks $\mathcal{X}$ constructed to be compatible with the OSWorld-Verified interaction protocol.
Each task $x\in\mathcal{X}$ contains a natural-language instruction, an initial environment state, and an automatic verifier.
The environment supports state reset, execution of mouse and keyboard actions in real desktop applications, and rule-based final-state evaluation.

At step $t$, the agent observes a multimodal observation $o_t$ and predicts an action from the interaction history $h_t=(o_1,a_1,\ldots,o_t)$.
A rollout defines as
\begin{equation}
\tau=(x,o_1,a_1,\ldots,o_T,a_T),
\end{equation}
and the verifier $V_x(\tau)\in\{0,1\}$ determines whether the final environment state satisfies the task goal.
The plain policy model is
\begin{equation}
\pi_{\mathrm{policy}}(a_t\mid h_t,x) \triangleq \pi_\theta(a_t\mid h_t,x),
\end{equation}
which is not conditioned on any skill.
Given a task-specific skill $s_x$, the skill-guided policy model can be expressed as
\begin{equation}
\pi_{\mathrm{skill}}(a_t\mid h_t,x,s_x) \triangleq \pi_\theta(a_t\mid h_t,x,s_x).
\end{equation}
Both policies are parameterized by the shared parameters $\theta$, operate over the same action space, and differ only in the conditioning context.
The skill acts as a training-time privileged recovery context that steers the current model toward a more informed recovery mode without changing its underlying interface.

Given a trajectory dataset $\mathcal{D}$, standard supervised fine-tuning optimizes the plain policy with the loss
\begin{equation}
\mathcal{L}_{\mathrm{SFT}}(\mathcal{D};\theta)
=
-\mathbb{E}_{\tau\sim\mathcal{D}}
\sum_{t=1}^{|\tau|}
\log \pi_\theta(y_t \mid h_t, x).
\end{equation}
Here, $y_t$ denotes the ground-truth action of the trajectory $\tau$.
This objective trains the policy to imitate the action labels in $\mathcal{D}$ and serves as the base training objective used throughout SGCD.

\section{Method}
\label{sec:method}
\subsection{Motivation}
\label{sec:Motivation}

Expert trajectories provide no effective supervision for policy-induced off-trajectory states~\cite{lauffer2025imitation,ross2011reduction}. Such states are not arbitrary perturbations: induced through the agent's closed-loop execution process, these states reflect the systematic biases of the learned policy. To inform the design of continuation supervision, we analyze failed rollouts from the plain policy (Fig.~\ref{fig:failure-analysis}) and identify two empirical properties: failures are strongly concentrated in early execution, and failed trajectories exhibit recurring error tendencies rather than isolated accidental mistakes.
These observations motivate \textbf{Skill-Guided Continuation Distillation} (SGCD), which first runs the plain policy to reach realistic off-trajectory states and then invokes the skill-guided policy to obtain successful continuations from these states.
As illustrated in Fig.~\ref{fig:method}, SGCD proceeds in four stages. In Stage I (Sec.~\ref{sec:task-trajectory-sampling}), we roll out the plain policy on different training tasks to collect successful and failed trajectories. In Stage II (Sec.~\ref{sec:skill-construction}), we summarize these trajectories and construct task-specific off-trajectory continuation skills. In Stage III (Sec.~\ref{sec:recovery-trajectory-construction}), the plain policy executes the first $k$ GUI actions on recoverable failure tasks to reach realistic off-trajectory states, after which the current GUI state is passed to the skill-conditioned policy to obtain verified successful continuations. In Stage IV (Sec.~\ref{sec:mixed-training}), we process the resulting continuations and incorporate these continuations with expert trajectories to optimize the deployment policy without skill prompts.

\subsection{Stage I: Task Trajectories Sampling}
\label{sec:task-trajectory-sampling}

The first stage samples the unassisted behavior of the current plain policy. For each training task $x\in\mathcal{X}$, we reset the environment and run the plain policy $M$ times, where $M$ denotes the number of rollouts sampled per task. The resulting set is
\begin{equation}
\mathcal{T}_{x,\mathrm{policy}} =
\left\{
\tau_{x,\mathrm{policy}}^{m} \sim \pi_{\mathrm{policy}}(\cdot \mid x)
\right\}_{m=1}^{M},
\label{eq:sample}
\end{equation}
where $\tau_{x,\mathrm{policy}}^{m}$ denotes the $m$-th complete trajectory collected by executing $\pi_{\mathrm{policy}}$ on task $x$.
Using the task verifier $V_x$, we partition all sampled rollouts of task $x$ into successful and failed trajectories:
\begin{equation}
\mathcal{D}_x^{+} = \{ \tau \in \mathcal{T}_{x,\mathrm{policy}} : V_x(\tau)=1 \},
\end{equation}
\begin{equation}
\mathcal{D}_x^{-} = \{ \tau \in \mathcal{T}_{x,\mathrm{policy}} : V_x(\tau)=0 \},
\end{equation}
The first stage deliberately collects rollouts from the plain policy rather than an external expert, because the resulting rollouts directly expose the off-trajectory states and recurring mistakes characteristic of the deployed agent's closed-loop behavior.
Successful rollouts provide feasible workflows and terminal evidence, while failed rollouts expose model-specific traps, target confusions, and premature termination patterns that are later summarized into off-trajectory continuation skills.

\subsection{Stage II: Skill Construction}
\label{sec:skill-construction}

We use Gemini-3-Pro~\cite{google2025gemini3} to transform task-level trajectory evidence into compact natural-language skills.
The skill format is motivated by the path-plural nature of GUI control.
Unlike math and code tasks, GUI tasks rarely admit a unique ground-truth solution trace. 
The same instruction for GUI agents may be completed through different menus, shortcuts, dialog states, file views, or interaction orders.
Directly conditioning on a successful path is therefore a brittle form of supervision.
It over-binds the current decision to a particular historical state sequence and preserves incidental low-level details such as menu order or scroll position. As a result, the model may learn to imitate a route rather than understand how the task can be recovered from an uncertain intermediate state.

SGCD represents a skill as a \emph{trajectory abstraction} rather than a trajectory replay. It summarizes trajectory-level evidence into off-trajectory continuation skills while filtering out incidental path details. We construct this abstraction from our failure analysis. As illustrated in Fig.~\ref{fig:failure-analysis}(b), failed GUI rollouts can be organized into four recurring failure families: \emph{Early Done}, where the model terminates before the task is actually complete; \emph{Fixation}, where it repeats ineffective actions without switching strategy; \emph{Hallucinated Affordance}, where it searches for unavailable menus, settings, or commands; and \emph{Scope Misjudgment}, where it selects an incorrect UI function or dialog despite understanding the broad instruction.

The skill schema is designed to guide successful continuations from off-trajectory states. It summarizes what should be achieved, which UI targets matter, what failure traps should be avoided, and how completion should be verified, while leaving the skill-conditioned policy free to choose a feasible continuation from the current GUI state. This abstraction is especially suitable for continuation construction, since the handoff state is produced online by the policy model and may not match any historical successful trace. Accordingly, each skill $s_x$ is indexed by task id and organized according to the schema in Tab.~\ref{tab:skill-schema}.

\begin{table}[t]
\centering
\small
\setlength{\tabcolsep}{4pt}
\renewcommand{\arraystretch}{1.15}
\begin{tabularx}{\linewidth}{@{}lX@{}}
\toprule
\textbf{Skill Field} & \textbf{Skill Role} \\
\midrule
\textbf{Continuation Plan} & Addresses \emph{Fixation} with high-level maneuvers and alternatives for switching away from stalled strategies. \\
\textbf{Critical Targets} & Addresses \emph{Scope~Misjudgment} by identifying task-relevant UI functions, files, widgets, cells, menus, or dialogs. \\
\textbf{Failure Traps} & Addresses \emph{Hallucinated~Affordance} by listing nonexistent affordances, misleading operations, and common dead ends. \\
\textbf{Success Criteria} & Addresses \emph{Early~Done} with observable completion checks specifying when termination is justified. \\
\bottomrule
\end{tabularx}
\caption{The structured schema of a task-specific off-trajectory continuation skill.}
\label{tab:skill-schema}
\end{table}

\subsection{Stage III: Off-Trajectory Continuation Construction}
\label{sec:recovery-trajectory-construction}
The third stage constructs successful continuations from recoverable failed tasks. A recoverable task is selected only if the plain policy fails while the skill-guided policy succeeds:
\begin{equation}
\mathcal{X}_{\mathrm{rec}} = \left\{ x\in\mathcal{X}_{\mathrm{fail}} : V_x(\tau_{x,\mathrm{skill}})=1 \right\},
\end{equation}
where $\mathcal{X}_{\mathrm{fail}}$ represents failed tasks of policy model $\pi_{\mathrm{policy}}$.
This filtering avoids spending sampling budget on tasks for which the skill-guided policy $\pi_{\mathrm{skill}}$ cannot yet find a successful solution, since such tasks are unlikely to yield verified successful continuations. Because SGCD is iterative, this recoverable task set is not fixed. As the plain policy improves in later rounds, additional tasks may become recoverable and enter subsequent off-trajectory continuation data construction.

As shown in Fig.~\ref{fig:failure-analysis}(a), failures are strongly concentrated in early execution. We therefore enumerate switch steps within this early window, i.e., $k\in\{1,\ldots,20\}$. By sweeping over multiple handoff depths, SGCD reduces hand-crafted bias in selecting off-trajectory states while preserving enough remaining horizon for the skill-conditioned policy to complete the task.

For each selected task $x\in\mathcal{X}_{\mathrm{rec}}$ and switch step $k$, we reset the environment and let the plain policy execute the first $k$ GUI actions. This produces a live policy-induced state with history
\begin{equation}
h_{k+1}^{p}=(o_1,a_1,\ldots,o_k,a_k,o_{k+1}) \sim\pi_{\mathrm{policy}}(\cdot\mid x).
\end{equation}
This action segment is not replayed from a stored trajectory. It is produced in the current environment, so the handoff state reflects a state that the plain policy actually reaches. Starting from this state, the skill-guided policy model continues execution with the task-specific skill $s_x$:
\begin{equation}
\hat{\tau}_{>k}
=
(\hat{a}_{k+1},\ldots,\hat{o}_{T},\hat{a}_{T})
\sim
\pi_{\mathrm{skill}}(\cdot\mid h_{k+1}^{p},x,s_x).
\end{equation}
The resulting successful continuation trajectory is
\begin{equation}
\hat{\tau}=(x,o_1,a_1,\ldots,
o_{k+1},\hat{a}_{k+1},\ldots,
\hat{o}_T,\hat{a}_T).
\end{equation}

Each spliced rollout is filtered by two complementary signals.
First, the executable verifier must accept the final state, $V_x(\hat{\tau})=1$.
Second, an LLM judge~\cite{zheng2023judging,lu2025agentrewardbench} inspects the task, observations, actions, and final outcome to remove accidental success, redundant loops, inconsistent reasoning, unsafe operations, and behavior unrelated to the instruction.
Only post-handoff continuations that pass both filters are added to the continuation dataset $\mathcal{D}_{\mathrm{cont}}$.

\subsection{Stage IV: Mixed Trajectories Training}
\label{sec:mixed-training}

The final stage trains the deployment policy on a mixture of original expert trajectories $\mathcal{D}_{\mathrm{exp}}$, verified successful policy trajectories $\mathcal{D}^{+}$, and verified successful continuations $\mathcal{D}_{\mathrm{cont}}$.
For each retained continuation trajectory $\hat{\tau}$, we discard the pre-handoff policy segment $(o_1,a_1^p,\ldots,o_k,a_k^p)$ from action supervision. This segment is used only as historical context and does not provide supervised action targets, since it may contain the policy behaviors that induced the off-trajectory state.

The continuation loss is
\begin{equation}
\mathcal{L}_{\mathrm{cont}}(\theta)
= -\mathbb{E}_{\hat{\tau} \sim\mathcal{D}_{\mathrm{cont}}}
\sum_{t=k+1}^{|\hat{\tau}|}
\log\pi_\theta(\hat{a}_{t}\mid \hat{h}_t, x),
\end{equation}
where each trajectory $\hat{\tau}$ comes with its source task $x$ and handoff step $k$. Here, $\hat{h}_t$ denotes the history used to predict the continuation action $\hat{a}_t$.
 
We combine standard supervised training on original expert trajectories and verified successful policy trajectories with continuation supervision:
\begin{equation}
\mathcal{L}(\theta)
=\mathcal{L}_{\mathrm{SFT}}(\mathcal{D}_{\mathrm{exp}}\cup\mathcal{D}^{+};\theta)
+\mathcal{L}_{\mathrm{cont}}(\mathcal{D}_{\mathrm{cont}};\theta).
\end{equation}
The objective remains a standard action-generation objective, but the mixed training data shifts supervision toward successful continuations from policy-induced off-trajectory states. After training, the deployment policy receives no skill. The skill context is used only as a data-synthesis scaffold.

SGCD achieves continual self-improvement through iterative application. After each round, the improved policy induces a new failure distribution, and previously unrecoverable tasks may become recoverable in subsequent rounds. Each iteration reruns trajectory sampling, updates skills, and regenerates verified successful continuations for training, progressively expanding continuation supervision as the policy's capability grows.

\definecolor{baserow}{RGB}{255,248,225}
\begin{table*}[!t]
\centering
\scriptsize
\setlength{\tabcolsep}{2pt}
\renewcommand{\arraystretch}{0.9}
\resizebox{0.9\textwidth}{!}{
\begin{tabular}{@{}lccccccc@{}}
\toprule
\multirow{2}{*}{\textbf{Method}} &
\multirow{2}{*}{\textbf{\#Params}} &
\multicolumn{6}{c}{\textbf{OSWorld-Verified}} \\
\cmidrule(lr){3-8}
& &
\textbf{Full$\uparrow$} &
\textbf{OS} &
\textbf{Office} &
\textbf{Daily} &
\textbf{Prof.} &
\textbf{Workflow} \\
\midrule

\rowcolor{gray!15}
\multicolumn{8}{c}{\textbf{General-Purpose Models}} \\
Seed1.8\cite{seed2026seed1} & -- & 61.9 & 16 & 81 & 29 & 58 & 39 \\
Kimi K2.5\cite{team2026kimi} & 1T & 63.3 & 17 & 81 & 29 & 59 & 43 \\
Claude Sonnet 4.6~\cite{anthropic2025claude4} & -- & 72.1 & 22 & 88 & 33 & 58 & 56 \\
GPT-5.5~\cite{openai2026gpt54} & -- & 78.7 & -- & -- & -- & -- & -- \\

\midrule
\rowcolor{gray!15}
\multicolumn{8}{c}{\textbf{GUI-Specialized Models}} \\
OpenCUA-7B~\cite{wang2026opencua} & 7B & 28.7 & 11 & 28 & 18 & 34 & 13 \\
UI-TARS-1.5-7B~\cite{qin2025ui} & 7B & 29.6 & 8 & 35 & 19 & 37 & 8 \\
TianXi-Action-7B & 7B & 30.4 & 8 & 34 & 19 & 43 & 5 \\
GUI-Owl-7B~\cite{ye2025mobile} & 7B & 32.1 & 12 & 29 & 19 & 47 & 9 \\
OpenCUA-32B~\cite{wang2026opencua} & 32B & 35.6 & 15 & 36 & 20 & 45 & 12 \\
DeepMiner-Mano-7B~\cite{fu2025mano} & 7B & 40.1 & 12 & 46 & 18 & 53 & 16 \\
DART-GUI-7B\cite{li2025dart} & 7B & 40.5 & 13 & 39 & 26 & 52 & 17 \\
OpenCUA-72B-preview~\cite{wang2026opencua} & 72B & 45.9 & 14 & 55 & 26 & 52 & 19 \\
UI-TARS-2~\cite{qin2025ui} & -- & 53.1 & 10 & 72 & 29 & 49 & 32 \\
DeepMiner-Mano-72B~\cite{fu2025mano} & 72B & 53.9 & 16 & 74 & 25 & 57 & 23 \\

\midrule
\rowcolor{gray!15}
\multicolumn{8}{c}{\textbf{Backbone-Matched Models}} \\

\multicolumn{8}{l}{\textit{Qwen3-VL-30B-A3B backbone}} \\
Lite-GUI-30B-A3B~\cite{wu2026litegui} & 30B-A3B & 22.7 & -- & -- & -- & -- & -- \\
\rowcolor{baserow}
Qwen3-VL-30B-A3B-Instruct$^\dagger$~\cite{bai2025qwen3} & 30B-A3B & 31.9 & 8 & 43 & 15 & 37 & 11 \\
Qwen3-VL-30B-A3B-Thinking$^\dagger$~\cite{bai2025qwen3} & 30B-A3B & 31.3 & 11 & 40 & 13 & 41 & 8 \\
Holo2-30B-A3B~\cite{hcompany2025holo2} & 30B-A3B & 37.4 & -- & -- & -- & -- & -- \\
\rowcolor{myrowblue}
\textbf{SGCD-30B-A3B} & \textbf{30B-A3B} & \textbf{58.4} & \textbf{18} & \textbf{78} & \textbf{17} & \textbf{55} & \textbf{43} \\

\cmidrule(lr){1-8}
\multicolumn{8}{l}{\textit{Qwen3-VL-8B backbone}} \\
\rowcolor{baserow}
Qwen3-VL-8B-Instruct$^\dagger$~\cite{bai2025qwen3} & 8B & 32.7 & 10 & 39 & 15 & 40 & 14 \\
Qwen3-VL-8B-Thinking$^\dagger$~\cite{bai2025qwen3} & 8B & 36.0 & 11 & 43 & 21 & 42 & 13 \\
Step-GUI-8B\cite{yan2025step} & 8B & 40.2 & 9 & 45 & 18 & 55 & 18 \\
EvoCUA-8B\cite{xue2026evocua} & 8B & 46.1 & 18 & 44 & 25 & 56 & 24 \\
GUI-Owl-1.5-8B-Instruct$^\dagger$\cite{xu2026mobile} & 8B & 52.4 & 18 & 59 & 19 & 65 & 28 \\
\rowcolor{myrowblue}
\textbf{SGCD-8B} & \textbf{8B} & \textbf{55.1} & \textbf{16} & \textbf{75} & \textbf{18} & \textbf{58} & \textbf{32} \\

\cmidrule(lr){1-8}
\multicolumn{8}{l}{\textit{STEP3-VL-10B backbone}} \\
\rowcolor{baserow}
STEP3-VL-10B$^\dagger$~\cite{huang2026step3} & 10B & 24.1 & 12 & 17 & 13 & 39 & 6 \\
\rowcolor{myrowblue}
\textbf{SGCD-10B} & \textbf{10B} & \textbf{53.2} & \textbf{13} & \textbf{69} & \textbf{19} & \textbf{59} & \textbf{32} \\

\bottomrule
\end{tabular}
}
\vspace{-0.6em}
\caption{Main results on OSWorld-Verified. We report overall OSWorld-Verified performance and category-level scores on different tasks. SGCD models are highlighted in \colorbox{myrowblue}{\textbf{blue}} and base models are in \colorbox{baserow}{yellow}. $^\dagger$ denotes results reproduced under our evaluation protocol. Officially reported full scores are 52.3 for GUI-Owl-1.5-8B-Instruct, 30.3 for Qwen3-VL-30B-A3B-Instruct, 30.6 for Qwen3-VL-30B-A3B-Thinking, and 33.9 for Qwen3-VL-8B variants.
}
\label{tab:osworld-main}
\end{table*}

\section{Experiments}
\label{sec:experiments}


\subsection{Experimental Setup}

\paragraph{Backbones.}


We apply SGCD to three vision-language GUI policy models with different scales and model families: Qwen3-VL-8B, Qwen3-VL-30B-A3B~\cite{bai2025qwen3}, and STEP3-VL-10B~\cite{huang2026step3}. This setting evaluates the effectiveness and generality of SGCD.

\paragraph{Training settings.}

All experiments are trained on 64 H100 GPUs. Following \cite{sun2025genesis,wu2025atlas}, we synthesize tasks over multiple real-world OS applications and construct the initial expert dataset by sampling trajectories with several advanced models, followed by filtering to retain high-quality successful executions.
Detailed training settings are provided in the appendix.

\paragraph{Baselines and metrics.}
We compare against commercial and GUI-specialized models, with the full list provided in Tab.~\ref{tab:osworld-main}.
Following \cite{li2025efficient,fu2025mano,agashe2025agent,gonzalez2025unreasonable}, we evaluate on the OSWorld-Verified \cite{xie2024osworld}, which reflects the real user experience.
We use end-to-end task success rate on OSWorld-Verified. 
To measure off-trajectory continuation ability, we define \emph{Continuation Success Rate}. We select tasks that the plain policy model fails. For each task, we first let the plain policy execute $k$ steps to induce an off-trajectory state, and then hand control to the evaluated model to complete the task from that state. Importantly, the evaluated model is not given any skill or guidance, so the metric measures whether the model has ability itself. 
Continuation Success Rate is the success rate of these tasks.





\subsection{Main Results on OSWorld-Verified}


Tab.~\ref{tab:osworld-main} reports the main comparison on OSWorld-Verified. We include overall performance and category-level scores on OS, Office, Daily, Professional, and Workflow tasks. The \emph{GUI-Specialized Models} block contains models specifically designed or trained for GUI tasks, while the \emph{Backbone-Matched Models} block compares methods that share the same base backbones as our models.
SGCD achieves state-of-the-art results among models with comparable parameter scales. In the backbone-matched comparison, SGCD substantially improves the original base models, with gains of over 20\% on the Qwen3-VL-8B backbone and over 25\% on the Qwen3-VL-30B-A3B backbone. These consistent improvements across different model families and scales demonstrate the effectiveness and generality of SGCD.
Overall, the results show that SGCD provides a strong and complementary training signal for GUI agents, beyond simply scaling the base model or relying on successful trajectory imitation.

\begin{figure*}[!t]
\centering

\begin{minipage}[t]{0.25\textwidth}
    \vspace{0pt}
    \centering
    \includegraphics[width=\linewidth]{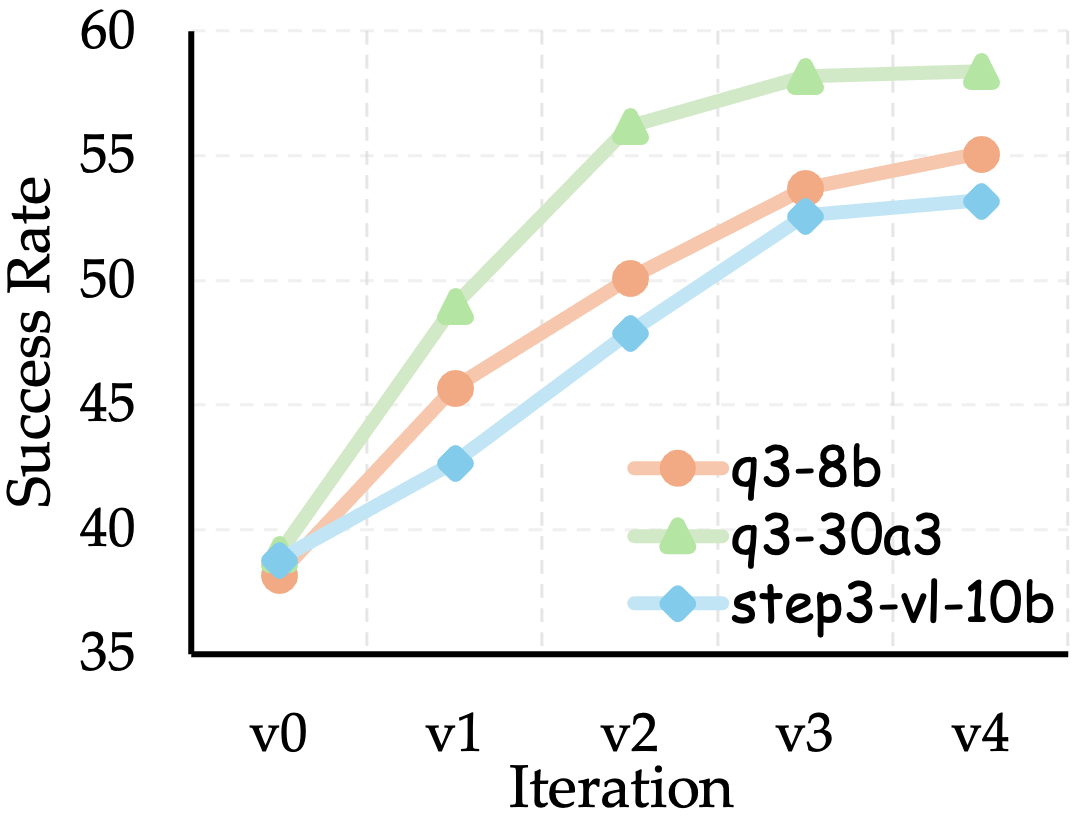}
    \vspace{-0.6em}
    \captionof{figure}{Iterative SGCD across training rounds.}
    \label{fig:iterative-recovery}
\end{minipage}
\hfill
\begin{minipage}[t]{0.25\textwidth}
    \vspace{0pt}
    \centering
    \includegraphics[width=\linewidth]{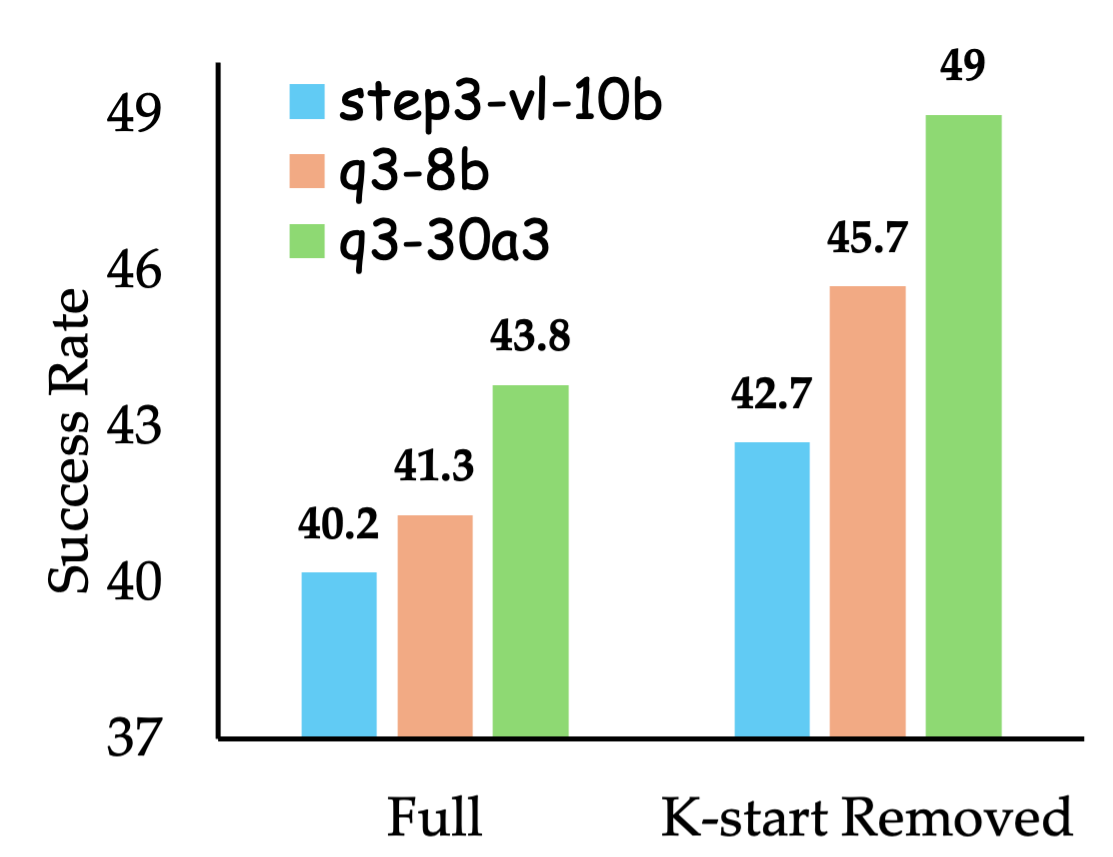}
    \vspace{-0.6em}
    \captionof{figure}{Effect of different starting rounds.}
    \label{fig:k-start}
\end{minipage}
\hfill
\begin{minipage}[t]{0.46\textwidth}
    \vspace{0pt}
    \centering
    \footnotesize
    \setlength{\tabcolsep}{1.8pt}
    \renewcommand{\arraystretch}{1.08}
    \begin{tabularx}{\linewidth}{@{}>{\raggedright\arraybackslash}Xcccccc@{}}
    \toprule
    \textbf{Skill Block} &
    \textbf{Full} &
    \textbf{OS} &
    \textbf{Office} &
    \textbf{Daily} &
    \textbf{Prof.} &
    \textbf{Work.} \\
    \midrule
    Without Skill & 39.9 & 10 & 51 & 19 & 45 & 19 \\
    CP & 41.3 & 12 & 48 & 15 & 59 & 15 \\
    CP + CT & 42.7 & 12 & 50 & 16 & 52 & 24 \\
    CP + CT + FT & 43.2 & 12 & 46 & 16 & 57 & 25 \\
    CP + CT + FT + SC & 45.7 & 12 & 52 & 16 & 61 & 24 \\
    \bottomrule
    \end{tabularx}
    \vspace{0.15em}
    \captionof{table}{Ablation on skill components. CP, CT, FT, and SC denote Continuation Plans, Critical Targets, Failure Traps, and Success Criteria.}
    \label{tab:skill-ablation}
\end{minipage}

\vspace{-0.5em}
\end{figure*}

\subsection{Off-Trajectory Continuation Ability}
We further evaluate whether SGCD trains the model to complete tasks from policy-induced off-trajectory states. For each task that the original policy fails, we first run the original policy for $k$ steps to induce an intermediate GUI state, and then let the evaluated model continue from that state without any skill prompt. This setting tests whether successful continuation ability is distilled into the plain policy itself.

As shown in Tab.~\ref{tab:recovery-rate}, SGCD attains the highest continuation success rate among open-weight models at both backbone scales, reaching $39.2\%$ on the Qwen3-VL-8B backbone and $50.3\%$ on the Qwen3-VL-30B-A3B backbone, which approaches Kimi K2.5.
This indicates that the skill-guided continuations are not merely useful during data synthesis, but are effectively distilled into the plain policy. The resulting model already learns how to complete tasks from off-trajectory states.

\begin{table}[t]
\centering
\small
\setlength{\tabcolsep}{5pt}
\renewcommand{\arraystretch}{1.08}
\begin{tabular}{@{}lcc@{}}
\toprule
\textbf{Method} & \textbf{\#Params} & \textbf{Cont. SR$\uparrow$} \\
\midrule
\rowcolor{gray!15}
\multicolumn{3}{c}{\textbf{General-Purpose Models}} \\
Kimi K2.5 & 1T & 56.3 \\
Claude Sonnet 4.6 & -- & 63.3 \\
\midrule
\rowcolor{gray!15}
\multicolumn{3}{c}{\textbf{Backbone-Matched Models}} \\
\multicolumn{3}{l}{\textit{Qwen3-VL-8B backbone}} \\
Qwen3-VL-8B-Instruct & 8B & 18.9 \\
Qwen3-VL-8B-Thinking & 8B & 27.2 \\
GUI-Owl-1.5-8B-Instruct & 8B & 28.2 \\
EvoCUA-8B & 8B & 32.7 \\
\rowcolor{myrowblue}
\textbf{SGCD-8B} & \textbf{8B} & \textbf{39.2} \\
\midrule
\multicolumn{3}{l}{\textit{Qwen3-VL-30B-A3B backbone}} \\
Qwen3-VL-30B-A3B-Instruct & 30B-A3B & 18.8 \\
Qwen3-VL-30B-A3B-Thinking & 30B-A3B & 23.3 \\
\rowcolor{myrowblue}
\textbf{SGCD-30B-A3B} & \textbf{30B-A3B} & \textbf{50.3} \\
\bottomrule
\end{tabular}
\caption{Continuation Success Rate (Cont. SR) comparison on the $142$ OSWorld-Verified tasks failed by the original plain policy.}
\label{tab:recovery-rate}
\end{table}


\subsection{Iterative Distillation}



SGCD is iterated because each trained policy induces a new distribution of off-trajectory states.
After one round, some previous failures become solvable, and new skill-solvable states emerge near the expanded capability boundary. We therefore repeat the pipeline of SGCD.

Fig.~\ref{fig:iterative-recovery} reports performance across SGCD training rounds. Performance improves consistently from v0 to later iterations, showing that repeated SGCD can further enhance the plain policy. 

\subsection{Ablation Studies}

We conduct all ablations on the v0-to-v1 stage of SGCD, where the largest improvement is observed.

\paragraph{Skill components.}

Tab.~\ref{tab:skill-ablation} studies the contribution of the off-trajectory continuation skill. Starting from no skill, we progressively add Continuation Plans, Critical Targets, Failure Traps, and Success Criteria. These components correspond to the dominant failure modes identified in our trajectory analysis: fixation, scope misjudgment, hallucinated affordance, and early done. The ablation tests whether the full skill works as a trajectory abstraction rather than a fixed path replay.

\paragraph{K-start supervision.}

SGCD uses only the post-handoff continuation for training. The pre-handoff segment is kept as context but excluded from action supervision. This design follows our motivation that off-trajectory states are more likely to produce erroneous actions. Therefore, the early plain-policy actions that lead to these states may contain noisy decisions rather than useful supervision. Fig.~\ref{fig:k-start} compares this design with training on the full trajectories. The results confirm that excluding the pre-handoff actions leads to better performance.


\FloatBarrier

\section{Conclusion}

We identify the off-trajectory supervision deficit as a fundamental limitation of behavior cloning on expert trajectories, where policy-induced states encountered during closed-loop execution receive no effective supervision signal.
To address this, we propose \textbf{Skill-Guided Continuation Distillation (SGCD)}, which leverages skill-guided rollouts to synthesize verified successful continuations from such states, supplying the supervision that expert trajectories cannot provide.
On OSWorld-Verified, SGCD improves the success rate of three base models from the low-30\% range to over 50\%, demonstrating its effectiveness and generality.

\section*{Limitations}


This study has several limitations. 
First, the current framework has limited effectiveness in acquiring successful continuations for difficult tasks. Improving the coverage and efficiency of continuation acquisition on such tasks remains an open direction for future work.
Second, SGCD currently instantiates off-trajectory states by re-executing the plain policy from scratch in the live environment for each task, which incurs substantial interaction overhead. Developing state-caching infrastructure to store and reuse intermediate GUI states directly could substantially reduce the number of environment steps required per iteration.

\bibliography{custom}

\clearpage
\appendix

\section{Additional Handoff-Depth Analysis}
\label{sec:handoff-depth-appendix}
To understand how the choice of handoff depth $k$ shapes the continuation dataset, we ablate the $k$ set used to construct $\mathcal{D}_{\mathrm{cont}}$ for Qwen3-VL-8B. All variants share the same skill, verifier, and training recipe described in Sec.~\ref{sec:mixed-training}; only the range of admissible handoff depths differs. Results are reported in Tab.~\ref{tab:k-ablation}.

\begin{table}[h]
\centering
\small
\begin{tabular}{lc}
\toprule
\textbf{\(k\) Set} & \textbf{Score} \\
\midrule
All Set \((1\text{--}20)\) & 45.7 \\
\(1\text{--}5\) & 42.9 \\
\(5\text{--}10\) & 43.2 \\
\(10\text{--}15\) & 41.0 \\
\(15\text{--}20\) & 41.8 \\
\bottomrule
\end{tabular}
\caption{Ablation on the handoff-depth set used for off-trajectory continuation construction.}
\label{tab:k-ablation}
\end{table}

\paragraph{Sweeping $k$ dominates any single bucket.}
Enumerating $k\in\{1,\ldots,20\}$ reaches $45.7$, outperforming every individual sub-range by $2.5$--$4.7$ points. This indicates that no single handoff depth is sufficient: different tasks expose their policy-induced off-trajectory states at different points along the rollout, and a fixed $k$ inevitably misses the off-trajectory state where the actionable mistake occurs. Sweeping the full window converts each failed task into multiple supervised continuations from distinct off-trajectory states, which both increases data volume and reduces the hand-crafted bias of any single handoff choice.

\paragraph{Performance degrades with deeper handoff.}
Within a fixed-width window, accuracy decreases monotonically as $k$ grows beyond $10$ ($43.2 \to 41.0 \to 41.8$). This matches the failure-distribution finding in Fig.~\ref{fig:failure-analysis}(a): plain-policy errors are concentrated in early execution, so an off-trajectory state induced after only a few steps lies closest to the failure mode that distillation needs to repair. Deeper handoff also leaves the skill-guided policy with a shorter remaining horizon to complete the task, which lowers the verifier pass rate and yields fewer high-quality continuations per task.
\section{Per-Handoff-Depth Continuation Success Rate}
\label{sec:per-k-recovery-appendix}

Tab.~\ref{tab:recovery-rate} in the main paper reports the aggregate Continuation Success Rate. Here we provide the full per-handoff-depth breakdown for ten systems on the same evaluation protocol. The pool consists of $142$ OSWorld-Verified tasks on which the plain Qwen3-VL-8B policy fails. For each task and each $k\!\in\!\{1,\ldots,20\}$, we let the plain policy execute the first $k$ steps to induce a policy-induced off-trajectory state and then hand control to the evaluated model, which must complete the task without any skill prompt. A trial is counted as successful when the verifier score is $\ge\!0.8$. The full $20\!\times\!10$ table is reported in Tab.~\ref{tab:per-k-recovery}.

\begin{table*}[h]
\centering
\scriptsize
\setlength{\tabcolsep}{3pt}
\renewcommand{\arraystretch}{1.05}
\begin{tabular}{rcccccccccc}
\toprule
& \multicolumn{2}{c}{\textbf{General-Purpose}} & \multicolumn{2}{c}{\textbf{SGCD (ours)}} & \multicolumn{2}{c}{\textbf{Open-Weight}} & \multicolumn{4}{c}{\textbf{Qwen3-VL Backbones}} \\
\cmidrule(lr){2-3}\cmidrule(lr){4-5}\cmidrule(lr){6-7}\cmidrule(lr){8-11}
$k$ & Claude S4.6 & Kimi K2.5 & SGCD-30A3B & SGCD-8B & EvoCUA-8B & GUI-Owl-1.5-8B & 8B-Inst & 8B-Thk & 30A3-Inst & 30A3-Thk \\
\midrule
1  & 80.28 & 61.97 & 59.15 & 42.25 & 28.87 & 27.46 & 12.68 & 20.42 & 13.38 &  9.15 \\
2  & 77.46 & 60.56 & 57.75 & 38.73 & 26.76 & 23.94 & 12.68 & 19.72 &  7.04 & 14.79 \\
3  & 78.17 & 66.20 & 57.04 & 36.62 & 32.39 & 26.76 & 11.97 & 25.35 & 11.97 & 15.49 \\
4  & 71.83 & 59.86 & 59.15 & 41.55 & 33.80 & 26.76 & 16.90 & 23.24 & 12.68 & 14.08 \\
5  & 71.83 & 58.45 & 59.86 & 37.32 & 30.28 & 26.06 & 14.08 & 23.94 & 13.38 & 17.61 \\
6  & 73.94 & 56.34 & 49.30 & 37.32 & 27.46 & 19.72 & 10.56 & 22.54 & 16.90 & 19.72 \\
7  & 71.83 & 56.34 & 48.59 & 36.62 & 37.32 & 24.65 & 16.90 & 26.76 & 12.68 & 22.54 \\
8  & 66.20 & 54.23 & 49.30 & 40.85 & 30.99 & 30.28 & 16.20 & 29.58 & 15.49 & 23.24 \\
9  & 76.76 & 59.86 & 52.82 & 47.89 & 33.10 & 23.24 & 17.61 & 25.35 & 15.49 & 23.94 \\
10 & 64.08 & 59.86 & 51.41 & 42.25 & 28.87 & 29.58 & 16.90 & 28.17 & 18.31 & 21.83 \\
11 & 66.90 & 61.27 & 44.37 & 38.73 & 26.76 & 27.46 & 21.83 & 32.39 & 19.01 & 23.94 \\
12 & 64.08 & 57.04 & 46.48 & 33.80 & 33.80 & 25.35 & 25.35 & 29.58 & 19.01 & 32.39 \\
13 & 54.93 & 57.75 & 45.07 & 38.03 & 35.21 & 30.99 & 19.01 & 28.87 & 21.83 & 26.76 \\
14 & 46.48 & 51.41 & 46.48 & 38.03 & 34.51 & 30.28 & 21.83 & 28.17 & 24.65 & 26.06 \\
15 & 47.18 & 51.41 & 47.18 & 35.21 & 41.55 & 30.99 & 23.94 & 30.99 & 25.35 & 30.28 \\
16 & 45.77 & 48.59 & 46.48 & 37.32 & 33.10 & 30.28 & 19.01 & 26.76 & 26.06 & 31.69 \\
17 & 55.63 & 51.41 & 47.18 & 40.14 & 34.51 & 34.51 & 21.83 & 33.80 & 23.24 & 28.17 \\
18 & 56.34 & 52.82 & 49.30 & 41.55 & 35.92 & 31.69 & 25.35 & 29.58 & 25.35 & 28.17 \\
19 & 50.00 & 52.82 & 47.18 & 42.25 & 33.80 & 33.80 & 27.46 & 28.17 & 26.06 & 26.76 \\
20 & 47.18 & 47.89 & 42.25 & 37.32 & 35.21 & 30.28 & 26.06 & 30.28 & 28.87 & 28.87 \\
\midrule
\textbf{Avg.} & \textbf{63.34} & \textbf{56.30} & \textbf{50.32} & \textbf{39.19} & \textbf{32.71} & \textbf{28.20} & \textbf{18.91} & \textbf{27.18} & \textbf{18.84} & \textbf{23.27} \\
\bottomrule
\end{tabular}
\caption{Per-handoff-depth Continuation Success Rate (\%) on the $142$ OSWorld-Verified tasks failed by the original plain policy. For each $k$, the plain policy executes $k$ steps before handing off to the evaluated model, which must complete the task without any skill prompt. Average row reports the unweighted mean over $k\!=\!1,\ldots,20$. 8B-Inst/Thk and 30A3-Inst/Thk denote the Qwen3-VL-8B-Instruct/Thinking and Qwen3-VL-30B-A3B-Instruct/Thinking backbones, respectively.}
\label{tab:per-k-recovery}
\end{table*}

\paragraph{SGCD beats backbone-matched baselines at every $k$.}
SGCD-30B-A3B averages $50.3\%$ vs.\ $28.2\%$ (GUI-Owl-1.5-8B-Instruct) and $27.2\%$ (Qwen3-VL-8B-Thinking), and the gap is consistent across the entire sweep, not driven by a few favorable handoff depths. SGCD-8B reaches $39.2\%$ vs.\ EvoCUA-8B's $32.7\%$ under the same backbone budget, with the gap widening on early-to-mid $k$ where the policy-induced off-trajectory state is most non-trivial.

\paragraph{Strong commercial models degrade fastest with $k$.}
Claude Sonnet 4.6 starts the highest at $k\!=\!1$ ($80.3\%$) but drops the most by $k\!=\!20$ ($47.2\%$), a $33$-point decline. Kimi K2.5 falls $14$ points over the same range. We attribute this to compounding prefix errors: the longer the plain Qwen3-VL-8B policy is allowed to act, the more its prefix accumulates mistakes that even strong general-purpose models cannot fully undo. SGCD-30B-A3B's range is much tighter ($42$--$60\%$), consistent with its training objective: continuation distillation explicitly supervises behavior from policy-induced off-trajectory states, so robustness across handoff depths is the property it is trained to acquire.

\paragraph{Weak baselines flatten or even improve with $k$.}
Models with very low $k\!=\!1$ Continuation Success Rate (Qwen3-VL-8B-Thinking\ at $20.4\%$; GUI-Owl-1.5-8B-Instruct at $27.5\%$) often perform better at moderate-to-deep $k$ than at $k\!=\!1$. This is because such models lack their own coherent open-ended plan; a moderately deep prefix from the plain policy effectively narrows the task and steers them toward the relevant UI surface, partially compensating for their weaker planning ability. SGCD-trained policies do not need this scaffolding -- they remain strong at small $k$, where the off-trajectory state is closest to the actual policy failure mode.

\section{Extended Related Work}
\label{sec:extended-related-work}

Vision-language GUI agents show strong promise in performing open-ended computer tasks~\cite{anthropic2024computeruse,anthropic2024computerusetool,openai2026gpt54,google2025gemini3,anthropic2025claude4,hurst2024gpt}. Standardized benchmarks across web, mobile, and desktop environments provide rigorous evaluation platforms for this line of research~\cite{deng2023mind2web,zhou2024webarena,zheng2024gpt,rawles2025androidworld,xie2024osworld}, and a growing body of GUI-specialized systems advances grounding, planning, action generation, and trajectory-level supervision~\cite{qin2025ui,wang2026opencua,yan2025step,xu2026mobile,hcompany2025holo2,wu2026litegui,xue2026evocua,ye2025mobile,cheng2024seeclick,gou2025navigating,lin2025showui}. SGCD is complementary to these advances: rather than proposing a new base model or benchmark, it targets the off-trajectory supervision deficit left by behavior cloning on expert trajectories.

Most GUI agents learn from successful demonstrations, synthetic trajectories, or state-action traces~\cite{deng2023mind2web,zheng2024gpt,qin2025ui,wang2026opencua,yan2025step,xu2026mobile,xue2026evocua}. Such supervision teaches grounding, action formatting, and interface conventions, but primarily covers expert-induced states. In closed-loop environments~\cite{zhou2024webarena,rawles2025androidworld,xie2024osworld}, local mistakes produce policy-induced off-trajectory states that are absent from successful traces. This is a direct manifestation of covariate shift in imitation learning~\cite{ross2011reduction,lauffer2025imitation}: the policy accumulates errors in states unseen during training, yet obtaining new supervision requires costly re-interaction with the live environment.

Self-improvement methods seek to address this by leveraging the agent's own experience as a training signal. Reflexion~\cite{shinn2023reflexion} and Self-Refine~\cite{madaan2023self} use inference-time verbal feedback to revise outputs. Recent GUI pipelines convert model-generated rollouts into supervision through filtered self-training~\cite{yan2025step,wu2026litegui}, sandbox-based reinforcement learning~\cite{lai2025computerrl}, experience-driven knowledge refinement~\cite{zhang2025ui,lin2026ui}, or policy-aligned experience assimilation~\cite{wang2026off}. SGCD follows this paradigm and specifically targets policy-induced off-trajectory states, synthesizing skill-guided continuation supervision from the states the current policy actually traverses.

Finally, structured guidance, memory, and experience retrieval improve agent planning~\cite{shinn2023reflexion,madaan2023self,agashe2025agent,qin2025ui,wang2026opencua,xue2026evocua}. Agent S uses experience-augmented hierarchical planning~\cite{agashe2025agent}, while OpenCUA, UI-TARS, and EvoCUA incorporate reflective conversion, memory-like capabilities, or self-correction~\cite{wang2026opencua,qin2025ui,xue2026evocua}. SGCD uses skills only as a training-time synthesis scaffold: Gemini-3-Pro abstracts successful and failed trajectories into Continuation Plans, Critical Targets, Failure Traps, and Success Criteria, enabling verified continuations without fixed-path replay or replacing the acting policy.
\section{Skill Extraction Prompt}
\label{sec:skill-prompt-appendix}

We use Gemini-3-Pro~\cite{google2025gemini3} to abstract trajectory evidence into structured off-trajectory continuation skills. For each task, the prompt is fed with the task id, all available successful reference trajectories sampled by the plain policy, and a matched set of failed reference trajectories. The model is instructed to (i) infer the stable good path from successful references, (ii) infer obstacles, traps, and bad paths from failed references, (iii) compress trajectories into reusable procedural knowledge rather than verbatim replays, and (iv) emit a JSON object with exactly four fields aligned with the failure families analyzed in Sec.~\ref{sec:skill-construction}: \emph{Continuation Plans}, \emph{Critical Targets}, \emph{Failure Traps}, and \emph{Success Criteria}. To suppress hallucinated guidance, alternative paths are only drawn from successful references, and fields without supporting evidence are left empty rather than filled in. The full prompt template is shown in Fig.~\ref{fig:skill-prompt}.

\begin{tcolorbox}[title=Skill Extraction Prompt (Gemini-3-Pro), colback=gray!5, colframe=gray!50, fonttitle=\bfseries\small, fontupper=\footnotesize, breakable]

\textbf{System Prompt}
\vspace{0.3em}

Task id: \texttt{\{task\_id\}}.

You are extracting a concise SOP (standard operating procedure) for one computer-use task. You will be given successful references and failed references for the same task.
\begin{itemize}[leftmargin=1.5em, nosep]
    \item Use \textbf{successful references} to infer the stable good path.
    \item Use \textbf{failed references} to infer obstacles, traps, and bad paths to avoid.
    \item Do not copy long trajectory text verbatim. Compress them into short, reusable procedural knowledge.
    \item Only include information supported by the provided references.
\end{itemize}

\textbf{Output Format}
\vspace{0.3em}

Return JSON only with exactly these fields:
\begin{verbatim}
{
  "Continuation Plans": string[],
  "Critical Targets":   string[],
  "Failure Traps":      string[],
  "Success Criteria":   string[]
}
\end{verbatim}

\textbf{Guidelines}
\begin{itemize}[leftmargin=1.5em, nosep]
    \item Keep each bullet short, concrete, and reusable.
    \item Prefer task-level decisions over click-by-click replay.
    \item Include wrong formulas / wrong pages / wrong menus in \texttt{Failure Traps} when supported by failed references.
    \item Alternative paths must only come from successful references; do not invent or generalize new alternatives.
    \item If uncertain, leave a field empty instead of hallucinating.
\end{itemize}

\vspace{0.5em}
\textbf{User Prompt}
\vspace{0.3em}

\textbf{Successful references}: \texttt{\{successful\_trajectories\}}\\
\textbf{Failed references}: \texttt{\{failed\_trajectories\}}

\end{tcolorbox}
\captionof{figure}{Skill-extraction prompt used with Gemini-3-Pro. Task-specific successful and failed reference trajectories are concatenated into the user-prompt slot, and the model is required to emit JSON with the four schema fields.}
\label{fig:skill-prompt}

\section{Skill Examples}
\label{sec:skill-examples-appendix}

To illustrate the structured output produced by the skill-extraction prompt, we show three representative skills covering different task families: a web browsing task on a Chinese auto-information portal (Fig.~\ref{fig:skill-example-miniev}), an office-document editing task on a word processor (Fig.~\ref{fig:skill-example-replace}), and a browser-settings task (Fig.~\ref{fig:skill-example-cookie}). Each skill contains four task-specific fields aligned with the failure families analyzed in Sec.~\ref{sec:skill-construction}: \emph{Continuation Plans} provide the abstracted good path, \emph{Critical Targets} list the salient UI anchors, \emph{Failure Traps} enumerate concrete mistakes observed in failed references, and \emph{Success Criteria} define the verifier-aligned terminal state.

\begin{tcolorbox}[title=Skill Example: Browse Wuling MINIEV owner community posts on Dongchedi, colback=gray!5, colframe=gray!50, fonttitle=\bfseries\small, fontupper=\footnotesize, breakable]

\textbf{Continuation Plans}
\begin{itemize}[leftmargin=1.5em, nosep]
    \item Find and click the ``Find Car'' or ``Select Car'' entry in the top navigation bar of the Dongchedi homepage (dongchedi.com).
    \item Type ``Wuling MINIEV'' in the search box and press Enter to search.
    \item On the search results page, click the ``Wuling MINIEV'' model card to enter the model detail page.
    \item On the model detail page, find and click the ``Owner Community'' or ``Community'' tab.
    \item Wait for the community feed to finish loading and confirm the page URL contains \texttt{/community/4499}.
\end{itemize}

\textbf{Critical Targets}
\begin{itemize}[leftmargin=1.5em, nosep]
    \item Dongchedi homepage search box.
    \item ``Wuling MINIEV'' search result card.
    \item ``Owner Community'' tab on the model detail page.
    \item Target URL: \texttt{dongchedi.com/community/4499}.
\end{itemize}

\textbf{Failure Traps}
\begin{itemize}[leftmargin=1.5em, nosep]
    \item Do not browse the homepage recommended content without searching for the specific model.
    \item Do not click ``Hongguang MINIEV'' or other similarly named model suggestions; confirm it is the ``MINIEV'' itself.
    \item Do not stay on the model specs or pricing page; you must navigate to the ``Owner Community'' tab.
    \item Do not click ad pop-ups or promotional banners on the page, as they navigate to irrelevant pages.
\end{itemize}

\textbf{Success Criteria}
\begin{itemize}[leftmargin=1.5em, nosep]
    \item The active tab URL is \texttt{dongchedi.com/community/4499}.
    \item The page displays the Wuling MINIEV owner community feed with a list of posts.
\end{itemize}

\end{tcolorbox}
\captionof{figure}{Skill extracted for a web browsing task on a Chinese auto-information portal. The Failure Traps field captures search-suggestion confusables (e.g., ``Hongguang MINIEV'' vs.\ ``Wuling MINIEV'') that repeatedly mislead the plain policy.}
\label{fig:skill-example-miniev}

\begin{tcolorbox}[title=Skill Example: Replace all question marks with exclamation marks, colback=gray!5, colframe=gray!50, fonttitle=\bfseries\small, fontupper=\footnotesize, breakable]

\textbf{Continuation Plans}
\begin{itemize}[leftmargin=1.5em, nosep]
    \item Press Ctrl+H to open the ``Find and Replace'' dialog.
    \item In the ``Search For'' input field, type ``?'' (question mark).
    \item Click the ``Replace With'' input field and type ``!'' (exclamation mark).
    \item Make sure the ``Regular expressions'' checkbox is NOT checked; otherwise ``?'' will be treated as a regex metacharacter.
    \item Click the ``Replace All'' button to perform the batch replacement.
    \item Confirm the replacement completion prompt (showing the number of replacements made).
    \item Press Escape to close the Find and Replace dialog.
    \item Press Ctrl+S to save the document.
\end{itemize}

\textbf{Critical Targets}
\begin{itemize}[leftmargin=1.5em, nosep]
    \item Ctrl+H to open the Find and Replace dialog.
    \item ``Search For'' input field (type ``?'').
    \item ``Replace With'' input field (type ``!'').
    \item ``Replace All'' button.
    \item Ctrl+S to save.
\end{itemize}

\textbf{Failure Traps}
\begin{itemize}[leftmargin=1.5em, nosep]
    \item Do not check the ``Regular expressions'' option, because ``?'' is a special character (quantifier) in regex and will cause the replacement to fail or match incorrectly.
    \item Do not check options under ``Other options'' that could affect matching, such as ``Current selection only''.
    \item Do not forget to check whether the ``Search For'' field has leftover text; clear it first before typing.
    \item Do not replace one by one (clicking ``Replace'' instead of ``Replace All''); this is extremely inefficient for documents with many occurrences.
\end{itemize}

\textbf{Success Criteria}
\begin{itemize}[leftmargin=1.5em, nosep]
    \item All original question marks in the document have been replaced with exclamation marks.
    \item All other content and formatting in the document remain unchanged.
    \item The document is saved.
\end{itemize}

\end{tcolorbox}
\captionof{figure}{Skill extracted for an office-document editing task. The Failure Traps field captures a non-obvious dialog-option pitfall (the ``Regular expressions'' checkbox interfering with the literal ``?'' match) that is hard to recover from once committed.}
\label{fig:skill-example-replace}

\begin{tcolorbox}[title=Skill Example: Enable blocking third-party cookies in Chrome, colback=gray!5, colframe=gray!50, fonttitle=\bfseries\small, fontupper=\footnotesize, breakable]

\textbf{Continuation Plans}
\begin{itemize}[leftmargin=1.5em, nosep]
    \item Type \texttt{chrome://settings/cookies} in the Chrome address bar and press Enter to navigate directly to the Cookie settings page.
    \item Alternatively, click the three-dot menu in the top-right corner $>$ Settings $>$ Privacy and Security $>$ Third-party cookies.
    \item On the Cookie settings page, find the ``Block third-party cookies'' option.
    \item Click the radio button or toggle for that option to set it to ``Block third-party cookies'' mode.
    \item Confirm the setting has taken effect (the option is selected or the toggle is in the on position).
\end{itemize}

\textbf{Critical Targets}
\begin{itemize}[leftmargin=1.5em, nosep]
    \item Chrome address bar (type \texttt{chrome://settings/cookies}).
    \item Or three-dot menu $>$ Settings $>$ Privacy and Security.
    \item ``Third-party cookies'' or ``Cookies and other site data'' setting.
    \item ``Block third-party cookies'' radio button/toggle.
\end{itemize}

\textbf{Failure Traps}
\begin{itemize}[leftmargin=1.5em, nosep]
    \item Do not select ``Block all cookies''; that will break many websites. Only block third-party cookies.
    \item Do not look for the Cookie option under ``Site settings''; it is under the main ``Privacy and Security'' category.
    \item Do not confuse ``Clear cookies'' with ``Block cookies''; they are two different operations.
    \item Do not close the browser to ``apply'' the settings; the change takes effect immediately.
\end{itemize}

\textbf{Success Criteria}
\begin{itemize}[leftmargin=1.5em, nosep]
    \item On Chrome's Cookie settings page, the ``Block third-party cookies'' option is selected/enabled.
    \item The setting persists after restarting Chrome.
\end{itemize}

\end{tcolorbox}
\captionof{figure}{Skill extracted for a browser-settings task. The Failure Traps field disambiguates closely related but semantically different options (``Block all cookies'' vs.\ ``Block third-party cookies'', ``Clear cookies'' vs.\ ``Block cookies'') that frequently cause plain-policy off-trajectory drift.}
\label{fig:skill-example-cookie}

\section{Training Task Construction}
\label{sec:data-construction-appendix}

We construct an OSWorld-compatible training pool $\mathcal{X}$ that is disjoint from the OSWorld-Verified~\cite{xie2024osworld} evaluation suite. Each task $x\in\mathcal{X}$ is a triple of (i) a natural-language instruction, (ii) a scripted initial environment state on the Ubuntu sandbox, and (iii) an automatic verifier returning $V_x(\tau)\in\{0,1\}$ from the post-rollout environment state. Whereas prior GUI training-data pipelines focus on trajectory synthesis~\cite{xu2025agenttrek,sun2025genesis,su2025learn,xue2026evocua} or grounding data~\cite{xie2026scaling}, our pipeline jointly produces the instruction, the initial-state assets, and the verifier from a per-application scenario specification, covers all ten desktop applications exercised by OSWorld-Verified, and is sized to support the trajectory-mixing recipe of Sec.~\ref{sec:mixed-training}.

\paragraph{Domain taxonomy.}
The ten target applications are organised into five coarse categories that serve as the unit of scenario design:
\begin{itemize}[leftmargin=1.5em, nosep]
    \item \textbf{OS} -- \texttt{os}: file manager, system settings, terminal, package and process management, and other OS-level operations.
    \item \textbf{Office} -- \texttt{libreoffice\_calc}, \texttt{libreoffice\_impress}, \texttt{libreoffice\_writer}: spreadsheet formatting, formulas, charts, and pivot tables; slide layouts, animations, and master styles; document editing, find-and-replace, styles, tables, and page setup.
    \item \textbf{Daily} -- \texttt{chrome}: web navigation, multi-tab search, form filling, bookmarks and history management, downloads, extensions, and browser-setting tasks.
    \item \textbf{Professional} -- \texttt{gimp}, \texttt{thunderbird}, \texttt{vlc}, \texttt{vs\_code}: image editing (layers, filters, color adjustments, selections, export); email composition, folder organisation, and mail-filter management; media playback control, playlists, subtitles, and conversion; code editing, search-and-replace, refactoring, extensions, debugging, and the integrated terminal.
    \item \textbf{Workflow} -- \texttt{multi\_apps}: cross-application tasks that compose two or more of the above, e.g., extracting data from a spreadsheet, rendering a chart in a presentation, and sending the result through email.
\end{itemize}
For each category we enumerate the modal usage scenarios and feature surfaces from official documentation and an LLM-driven web research pass over real-world usage patterns. The resulting per-application feature taxonomy is used purely as a sampling scaffold for instruction synthesis and is never fed to the trained policy.

\paragraph{Instruction synthesis and human filtering.}
Conditioned on the per-application feature taxonomy, an LLM proposes candidate natural-language instructions under three constraints:
(i) \textit{realistic} -- the task should plausibly arise from real user workflows rather than being a synthetic UI exercise;
(ii) \textit{concrete and verifiable} -- task completion must be decidable by inspecting the post-rollout environment;
(iii) \textit{deterministic verifier-relevant outcome} -- successful completion must be decidable from a well-defined set of terminal-state properties, even when multiple interaction paths or irrelevant UI configurations are possible..
A subsequent human pass de-duplicates near-identical instructions, removes ambiguous or under-specified phrasings, and rejects tasks whose terminal state cannot be operationalised without subjective judgement.

\paragraph{Initial-state assets and automatic verifiers.}
For every accepted instruction we manually prepare the assets required to instantiate the task on a fresh VM (e.g., the working spreadsheet, source document, reference image, mailbox snapshot, or browser profile), together with the reference file(s) used by the verifier. The initial state is realised by a per-task setup script that places these assets at the canonical paths expected by the OSWorld-Verified contract. The verifier itself follows one of two patterns according to the task family:
\begin{itemize}[leftmargin=1.5em, nosep]
    \item \textbf{State-extraction verifiers.} For tasks whose goal is to navigate the system or an application into a specific configuration (e.g., reach a particular Chrome setting page, select a specific track in VLC, switch a VS~Code panel), the verifier programmatically extracts the final application state through the application's introspection API or accessibility tree and asserts equality with the expected state.
    \item \textbf{File-match verifiers.} For tasks whose goal is to produce or modify a file (e.g., editing a Calc workbook, exporting a GIMP image, replacing strings in a Writer document), the verifier loads the post-rollout artifact and compares it against the reference file using property-level assertions (cell value/format, image content/metadata, document text/styles) rather than byte-level equality.
\end{itemize}

\paragraph{Difficulty binning.}
We label every task with the number of GUI primitives that an expert demonstrator needs to complete it, and bin tasks into three difficulty levels: \textit{easy} (1--5 steps), \textit{medium} (5--15 steps), and \textit{hard} (15--100 steps). The three bins are sampled to be approximately uniformly distributed within each application, so that the training pool exercises both short skill-grounding tasks and long-horizon planning.

\paragraph{Manual solvability check.}
Before a task enters the training pool, a human annotator executes it end-to-end on a fresh sandbox VM and confirms that the prepared initial state, the assets, and the automatic verifier are mutually consistent: the instruction is unambiguously actionable, the verifier returns $V_x(\tau)=1$ on the human-completed trajectory, and no required asset or UI surface is missing. Tasks that fail this manual pass -- because of an under-specified instruction, an unreachable goal state, or a verifier that disagrees with a correct human execution -- are sent back for asset, instruction, or verifier revision and re-tested, or dropped if they cannot be repaired. This step closes the gap between syntactic verifier acceptance and end-to-end solvability, and guarantees that every released task in $\mathcal{X}$ has at least one verified human solution.

\paragraph{Final corpus.}
The resulting training pool contains $1{,}432$ verified tasks spanning the ten domains; per-application counts are reported in Tab.~\ref{tab:training-data-counts}. The corpus is intentionally weighted toward the \textsc{Workflow} category, since multi-application tasks expose the long-horizon off-trajectory states that SGCD is designed to repair.

\begin{table}[h]
\centering
\small
\setlength{\tabcolsep}{6pt}
\renewcommand{\arraystretch}{1.05}
\begin{tabular}{@{}llr@{}}
\toprule
\textbf{Category} & \textbf{Application} & \textbf{\#Tasks} \\
\midrule
OS & \texttt{os} & 88 \\
\midrule
\multirow{3}{*}{Office}
 & \texttt{libreoffice\_calc}    & 189 \\
 & \texttt{libreoffice\_impress} & 179 \\
 & \texttt{libreoffice\_writer}  &  90 \\
\midrule
Daily & \texttt{chrome} & 188 \\
\midrule
\multirow{4}{*}{Professional}
 & \texttt{gimp}        & 86 \\
 & \texttt{thunderbird} & 53 \\
 & \texttt{vlc}         & 58 \\
 & \texttt{vs\_code}    & 90 \\
\midrule
Workflow & \texttt{multi\_apps} & 411 \\
\midrule
\textbf{Total} & & \textbf{1{,}432} \\
\bottomrule
\end{tabular}
\caption{Per-application task counts in the constructed training pool $\mathcal{X}$. Tasks are organised into five coarse categories (OS, Office, Daily, Professional, Workflow) used as the unit of scenario and feature analysis during instruction synthesis.}
\label{tab:training-data-counts}
\end{table}

\section{Algorithm Overview}
\label{sec:algorithm-appendix}

Algorithm~\ref{alg:sgcd} summarizes the full Skill-Guided Continuation Distillation pipeline. Each iteration samples plain-policy rollouts, extracts a task-specific skill from successful and failed evidence, instantiates policy-induced off-trajectory states with a $k$-step plain-policy prefix, hands the state off to the skill-guided policy to obtain a verified successful continuation, and distills the post-handoff portion back into the plain policy without skill prompts. As the policy improves across iterations, more failed tasks become recoverable and enter subsequent continuation construction.

\begin{algorithm}[h]
\small
\caption{Skill-Guided Continuation Distillation}
\label{alg:sgcd}
\begin{algorithmic}[1]
\State \textbf{Input:} tasks $\mathcal{X}$, expert trajectories $\mathcal{D}_{\mathrm{exp}}$, model $\pi_\theta$, skill constructor $G$, number of iterations $R$, handoff range $K_{\max}$
\For{iteration $r=1$ to $R$}
    \State Initialize $\mathcal{D}^{+}\leftarrow\emptyset$, $\mathcal{D}_{\mathrm{cont}}\leftarrow\emptyset$, $\mathcal{X}_{\mathrm{fail}}\leftarrow\emptyset$
    \For{task $x\in\mathcal{X}$}
        \State Sample policy trajectories $\mathcal{T}_{x,\mathrm{policy}}$ using $\pi_{\mathrm{policy}}$
        \State Add verified successes to $\mathcal{D}^{+}$ and failed tasks to $\mathcal{X}_{\mathrm{fail}}$
    \EndFor
    \For{task $x\in\mathcal{X}_{\mathrm{fail}}$}
        \State Construct task skill $s_x\leftarrow G(\mathcal{T}_{x,\mathrm{policy}})$ from success and failure evidence
        \State Sample $\tau_{x,\mathrm{skill}}\sim\pi_{\mathrm{skill}}(\cdot\mid x,s_x)$
        \If{$V_x(\tau_{x,\mathrm{skill}})=0$} \textbf{continue} \Comment{not recoverable yet}
        \EndIf
        \For{$k=1$ to $K_{\max}$}
            \State Run $\pi_{\mathrm{policy}}$ online for $k$ steps to reach $h_{k+1}^{p}$
            \State Hand off to $\pi_{\mathrm{skill}}(\cdot\mid h_{k+1}^{p},x,s_x)$ for the remaining horizon
            \If{spliced rollout $\hat{\tau}$ passes verifier $V_x$ and LLM quality judge}
                \State Add only post-handoff examples $(\hat{h}_t,\hat{a}_t)$ for $t>k$ to $\mathcal{D}_{\mathrm{cont}}$
            \EndIf
        \EndFor
    \EndFor
    \State Train $\pi_\theta$ on $\mathcal{D}_{\mathrm{exp}}\cup\mathcal{D}^{+}\cup\mathcal{D}_{\mathrm{cont}}$ without skill prompts
\EndFor
\State \textbf{Output:} deployment policy $\pi_\theta$ (no skill at inference)
\end{algorithmic}
\end{algorithm}

\section{Training Details}
\label{sec:training-details-appendix}

All three backbones are trained on 64 H100 GPUs (8 nodes $\times$ 8 GPUs) with the same trajectory mixture described in Sec.~\ref{sec:mixed-training}: original expert trajectories $\mathcal{D}_{\mathrm{exp}}$, verified successful policy trajectories $\mathcal{D}^{+}$, and verified post-handoff continuations $\mathcal{D}_{\mathrm{cont}}$. The Vision Transformer (ViT) and the vision--language aligner are frozen throughout, and only the language tower is fine-tuned. Training uses bfloat16 with the AdamW optimizer.

\paragraph{Qwen3-VL-8B and Qwen3-VL-30B-A3B.} These two backbones share the same training recipe under the \textsc{Swift} (Megatron-SFT) framework~\cite{zhao2025swift}. We use tensor-parallel size 2 with sequence parallel enabled, micro batch size 2 and global batch size 512, and a maximum sequence length of $10{,}240$. Activation recomputation is set to full with the uniform method (one layer per recompute block) and ViT gradient checkpointing is enabled. Optimization uses learning rate $1\!\times\!10^{-5}$ with a $1\!\times\!10^{-6}$ minimum, a warmup fraction of $0.05$, fused cross-entropy loss, and last-round loss masking so that supervision is applied only to the policy-predicted tokens of each turn. We train for $2$ epochs over the mixed dataset with packing enabled to reduce padding overhead. The pre-handoff plain-policy actions in $\mathcal{D}_{\mathrm{cont}}$ are kept as context but masked out of the action loss (K-start; Sec.~\ref{sec:mixed-training}).

\paragraph{STEP3-VL-10B.} This backbone is trained with the \textsc{Steptron} framework using a long-context configuration tailored to its YARN-extended $128\mathrm{k}$ position range. The model uses tensor-parallel size 8, pipeline-parallel size 1, sequence parallel, full activation recomputation, and freezes the ViT encoder. The training packs sequences to a global sequence length of $128 \times 1024 = 131{,}072$ tokens, with micro batch size 1 and global batch size 32 over $3{,}873$ iterations (approximately $2$ epochs of the trajectory mixture). Optimization uses AdamW ($\beta_1=0.9$, $\beta_2=0.95$, $\epsilon=10^{-8}$) with gradient clipping at $1.0$, a cosine schedule with peak learning rate $1\!\times\!10^{-4}$, minimum learning rate $1\!\times\!10^{-5}$, and a $200$-iteration linear warmup. Weight decay is set to $0$, and bf16 is used without fp16 loss scaling. Key training hyperparameters for the three backbones are summarized in Tab.~\ref{tab:training-hparams}.

\begin{table}[h]
\centering
\scriptsize
\setlength{\tabcolsep}{3.5pt}
\renewcommand{\arraystretch}{1.05}
\begin{tabular}{@{}lcc@{}}
\toprule
\textbf{Setting} & \textbf{Qwen3-VL-8B / 30B-A3B} & \textbf{STEP3-VL-10B} \\
\midrule
Framework & Swift & Steptron \\
Hardware & 64$\times$H100 & 64$\times$H100 \\
TP / PP & 2 / 1 & 8 / 1 \\
Seq.\ parallel & \checkmark & \checkmark \\
Micro / global bsz. & 2 / 512 & 1 / 32 \\
Seq.\ length & 10{,}240 & 131{,}072 \\
Optimizer & AdamW & AdamW \\
LR (peak/min) & 1e-5 / 1e-6 & 1e-4 / 1e-5 \\
LR schedule & cosine, 5\% warmup & cosine, 200 iter warmup \\
Train length & 2 epochs & 3{,}873 iter ($\approx$2 ep.) \\
Precision & bf16 & bf16 \\
Freeze ViT / aligner & \checkmark / \checkmark & \checkmark / \checkmark \\
\bottomrule
\end{tabular}
\caption{Training hyperparameters. Qwen3-VL-8B and Qwen3-VL-30B-A3B share the same Swift recipe; STEP3-VL-10B uses the Steptron long-context configuration with its YARN-extended $128\mathrm{k}$ position range.}
\label{tab:training-hparams}
\end{table}

\section{Evaluation Setup}
\label{sec:eval-setup-appendix}

\paragraph{Environment.} We follow the OSWorld-Verified~\cite{xie2024osworld} interaction protocol and use the OSWorld-Verified~\cite{xie2024osworld} task suite for evaluation. The agent acts in an Ubuntu desktop sandbox with a fixed screen size of $1920 \times 1080$, predicting one action per step from raw screenshots. The action space is \texttt{pyautogui} for mouse and keyboard primitives, augmented with two control functions: \texttt{computer.wait} for synchronous pauses (e.g., during installation or long computation) and \texttt{computer.terminate} for declaring task completion or failure. Coordinates are predicted in normalized form (relative to image size) and projected back to absolute pixels before execution.

\paragraph{Inference parameters.} For all backbones we use the same sampling configuration: \texttt{max\_tokens}~$=4096$, \texttt{top\_p}~$=0.95$, \texttt{temperature}~$=1.0$. Each task is run with at most $100$ environment steps; the agent is forced to terminate with failure once the budget is exhausted. At the API level, we allow up to $10$ outer retries for parse-level errors and up to $20$ HTTP-level retries per outer call, with a $1{,}200$-second request timeout.

\paragraph{Message construction.}
Each step the agent assembles an OpenAI-style chat-completion payload with the following structure:
\begin{itemize}
    \item A \texttt{system} message containing the GUI-agent system prompt (Fig.~\ref{fig:system-prompt}). The prompt fixes the response format (\textit{Thought}, \textit{Action}, \textit{Code}), the allowed action space, and a hard-coded sandbox password placeholder that some tasks require.
    \item Interleaved \texttt{user}/\texttt{assistant} turns for previous steps. To control context length, only the most recent $H=3$ historical screenshots are included as \texttt{image\_url} messages, each paired with the corresponding \texttt{assistant} response containing the past \textit{Thought} and \textit{Action}. Steps older than the image-history window are kept as text-only assistant context to preserve the planning chain without their screenshots.
    \item A final \texttt{user} turn containing the current screenshot (base64-encoded PNG) and the task instruction rendered through the \texttt{INSTRUCTION\_TEMPLATE} (``Please generate the next move according to the screenshot, task instruction and previous steps'').
\end{itemize}
For thinking-capable backbones, the assistant turns are rendered with explicit \verb|[think] ... [/think]| segments so that the historical reasoning is reusable; otherwise, the standard \texttt{\#\# Thought / \#\# Action} markdown layout is used. After each model call, the response is parsed for the \texttt{Action} block and the trailing fenced code block (\texttt{python} or \texttt{code}); a parse failure triggers a retry with reduced temperature.

\begin{tcolorbox}[title=Evaluation System Prompt, colback=gray!5, colframe=gray!50, fonttitle=\bfseries\small, fontupper=\footnotesize, breakable]
You are an agent that operates a computer GUI. For each step, you are given the task goal, the latest screenshot, and the history of earlier actions. Choose the single best next move that advances the task based on the current screen. Keep your action consistent with what is visible, and use the action history to avoid repeating mistakes. The password of the computer is \texttt{\{password\}}.

\vspace{0.3em}
For each step, provide your response in this format:

\begin{verbatim}
## Thought: {thought}
## Action:  {action}
## Code:    {code}
\end{verbatim}
Code rules:

- Return exactly one executable code block.

- The code block must contain either:

  1) pyautogui code for the next GUI action, or
  
  2) one built-in control call:
  
     - computer.wait(): wait 20 seconds if the page, app, or system needs time
     
     - computer.terminate(status=..., answer=...): end the task when it is finished or cannot be completed
     
- If you use computer.terminate, status must be "success" or "failure".

- Include answer only when there is a final result to report.

\end{tcolorbox}
\captionof{figure}{System prompt used at evaluation. Thinking-capable backbones receive a variant in which the \textit{Thought} field is replaced by an explicit reasoning segment wrapped in \texttt{[think]}\,/\,\texttt{[/think]} tags.}
\label{fig:system-prompt}

\end{document}